\documentclass[journal]{IEEEtran}
\usepackage[dvips]{epsfig}
\usepackage[utf8]{inputenc}
\usepackage{amssymb}
\usepackage{amsmath}
\usepackage{amsthm}
\usepackage{nccmath}
\usepackage{graphicx}
\usepackage{color}
\usepackage{mathtools}
\usepackage{algpseudocode}
\usepackage{algorithm}
\usepackage{cancel}
\usepackage{subfig}

\theoremstyle{definition}
\newtheorem{lemma}{Lemma}
\newtheorem{definition}{Definition}
\ifCLASSINFOpdf
\else
\fi
\begin{document}
%
% paper title
% Titles are generally capitalized except for words such as a, an, and, as,
% at, but, by, for, in, nor, of, on, or, the, to and up, which are usually
% not capitalized unless they are the first or last word of the title.
% Linebreaks \\ can be used within to get better formatting as desired.
% Do not put math or special symbols in the title.
\title{Parameterization of state duration in Hidden semi-Markov Models: an application in electrocardiography}
%
%
% author names and IEEE memberships
% note positions of commas and nonbreaking spaces ( ~ ) LaTeX will not break
% a structure at a ~ so this keeps an author's name from being broken across
% two lines.
% use \thanks{} to gain access to the first footnote area
% a separate \thanks must be used for each paragraph as LaTeX2e's \thanks
% was not built to handle multiple paragraphs
%

\author{Adrián~Pérez~Herrero, Paulo~F\'elix~Lamas, Jes\'us~Mar\'ia~Rodr\'iguez~Presedo % <-this % stops a space
\thanks{*Adrián Pérez Herrero is a PhD researcher from Centro 
Singular de Investigaci\'on en Tecnolox\'ias Intelixentes (CiTIUS), 
University of Santiago de Compostela, 15782 ESPAÑA.}% <-this % stops a space
\thanks{This work has been funded by Spanish Minister of Science and Innovation as part of the project RTI2018-099646-B100.}% <-this % stops a space
\thanks{}}

\maketitle

% As a general rule, do not put math, special symbols or citations
% in the abstract or keywords.
\begin{abstract}
This work aims at providing a new model for time series classification based on learning from just one example. We assume that time series can be well characterized as a parametric random process, a sort of Hidden semi-Markov Model representing a sequence of regression models with variable duration. We introduce a parametric stochastic model for time series pattern recognition and provide a maximum-likelihood estimation of its parameters. Particularly, we are interested in examining two different representations for state duration: i) a discrete density distribution requiring an estimate for each possible duration; and ii) a parametric family of continuous density functions, here the Gamma distribution, with just two parameters to estimate. An application on heartbeat classification reveals the main strengths and weaknesses of each alternative.
\end{abstract}

% Note that keywords are not normally used for peerreview papers.

% For peer review papers, you can put extra information on the cover
% page as needed:
% \ifCLASSOPTIONpeerreview
% \begin{center} \bfseries EDICS Category: 3-BBND \end{center}
% \fi
%
% For peerreview papers, this IEEEtran command inserts a page break and
% creates the second title. It will be ignored for other modes.
\IEEEpeerreviewmaketitle

\section{Introduction}

\IEEEPARstart{T}{ime} series classification has long been a challenge for the scientific community of machine learning, with an increasing demand for applications to speech recognition, signature verification, audio recognition and detection of cardiovascular diseases, amongst others. The modern approach to this problem emphasizes the importance of achieving results comparable to and even surpassing human performance. Generally, it is common knowledge that such a goal requires models with a large number of parameters and, consequently, learning requires a larger number of training examples.

A paradigm shift is proposed in \cite{FeiFei06} which, taking inspiration from human learning, addresses the challenge of obtaining information about a class from just one, or a handful, of examples. The key insight is that, once a few classes have been learned the hard way, some information may be abstracted from that process to make learning new classes more efficient. This intuition has motivated an emerging interest in a new area of research called one-shot learning, with applications in image recognition \cite{FeiFei06,Koch15,Lake11}, speech recognition \cite{Lake14} and language modelling \cite{Vinyals16}.

In this paper we propose a new model for learning time series patterns from just one example. The model is an extension of a Hidden semi-Markov Model \cite{Rabiner89,Murphy02,Yu10}, where each observable distribution is represented as a linear combination of fixed non linear functions from an orthonormal basis. Thus, it is intended to abstract or idealize a time series pattern as a sequence of regression probability distributions. As a paramount ability of the model, an unsupervised segmentation of the time series is required. To this aim, a duration probability distribution associated with each latent state is introduced. Two different options are explored for modelling each duration probability distribution: i) a non parametric discrete probability distribution; and ii) a parametric Gamma probability distribution. They will be compared in terms of expressiveness and computational efficiency. 

The viability of the model for one-shot learning under this two options is tested on real data from the electrocardiograph domain, with a set of experiments to validate its abstraction and recognition ability, and its computational performance.

The rest of this paper is outlined as follows. Section II formally defines the model. Section III shows an efficient method for computing the likelihood of a new time series from a given model. Section IV describes a method for training the model as an estimation procedure from a single time series. Then, a set of experiments for testing the viability of the model is presented in Section V. Finally, Section VI provides some conclusions and offers an outlook for future research.

\section {Definitions}
Let us suppose that a certain system is at any time in one of N distinct non-observable states $S_1, S_2, ..., S_N$. Consider an $N\times N$ {\em state transition} matrix $\mathbf{A}=[a_{ij}]$, where
\begin{displaymath} 
a_{ij}=P(q_{s+1}=S_j | q_s=S_i), \qquad 1\leq i,j \leq N,
\end{displaymath} 
\noindent that is, $a_{ij}$ is the probability of reaching state $S_j$ given current state $S_i$.  Consider also an {\em initial state distribution} vector 
$\boldsymbol{\pi}=(\pi_1,\pi_2,...,\pi_N)$, where
\begin{displaymath} 
\pi_i=P(q_1=S_i),\qquad 1\leq i \leq N,
\end{displaymath} 
\noindent that is, $\pi_i$ is the probability of beginning in state $S_i$. Both $\mathbf{A}$ and $\boldsymbol{\pi}$ specify an stochastic process 
called first-order N-state Markov chain.

Consider a {\em duration probability density} $p_i(d)$, $d\in \mathbb{Z}^+$, associated with state $S_i$, where
\begin{displaymath}
 p_i(d) = P(d|q_s=S_i),\qquad 1\leq i \leq N,
\end{displaymath}
When the system enters state $S_i$, a transition to another state is made only after an appropriate time interval, constrained by the duration density \cite{Rabiner89}. Accordingly, we enforce that $a_{ii}=0$, $1\leq i \leq N$.  We denote by $\boldsymbol{\delta}=(d_1,...,d_N)$ the set of parameters specifying the set of duration probability densities.

Let us now suppose that our system can only be observed through a set of stochastic processes that produce a time series $v_t=(t.T_p)$, $t\in \mathbb{N}$ by a sampling procedure, being $T_p$ the sampling period. We assume that the system has evolved through $q_1,q_2,...,q_s$ states during the first $t$ samples of the time series, with a duration of each state of $d_1,d_2,...,d_s$. We refer to the index $s$ as the state index; we call the sequence $(q_s)_{s=1}^S$ the state sequence and the sequence $(d_s)_{s=1}^S$ the duration sequence. For the sake of simplicity, we assume that any duration specifies a number of samples from the time series. Thus, at any time $t$ the system is in a state $q_s$ such that $\sum ^{s-1} _{r=0} d_r < t \leq \sum ^s _{r=1} d_r$ where $d_0=0$. 

Consider an {\em observable probability distribution} $b_i(v_t)$, associated with state $S_i$, where 
\begin{displaymath}
 b_i(v_t)=P(v_t|q_s=S_i,\boldsymbol{\theta}_i),
\end{displaymath}
\noindent being $\boldsymbol{\theta}_i$ a set of parameters governing the density function in the state $S_i$. We assume that the observable time series is given by 
$v_t=y(t)+\epsilon$, where $y(t)$ is a deterministic function and $\epsilon$ represents zero mean additive Gaussian noise with precision $\beta$. Thus, $b_i(v_t)$ can then be written as
\begin{equation}
 b_i(v_t)=\mathcal{N}(v_t|y_i(t),\beta_i^{-1}).
 \label{eq:emission}
\end{equation}
We assume that the deterministic function is modeled by a linear combination of fixed nonlinear functions, so that 
\begin{displaymath}
 y_i(t)=\sum ^{M}_{j=0} w_{ij}\phi_j(t)=\boldsymbol{w}_i^T\boldsymbol{\phi}(t),
\end{displaymath}
\noindent where $\boldsymbol{w}_i=(w_{i0},...,w_{iM})^T$ and $\boldsymbol{\phi}=(\phi_{0},...,\phi_{M})^T$. Parameters $w_{i0}$ can introduce an offset in the data, by setting the basis function $\phi_0(t)=1$. In this way, a different linear model can be indexed by any state $S_i$. There are many possible choices for the basis functions: polynomial, orthogonal polynomial (Hermite, Legendre, ...), Gaussian, sigmoidal, wavelet, etc. Thus, for any state $S_i$ a linear model can be specified by a vector of parameters $\boldsymbol{\theta}_i =(\boldsymbol{w}_i^T,\beta_i^{-1})$ and a set of basis functions $\boldsymbol{\phi}(t)$. For the sake of simplicity we will use the same set of basis functions for modeling all the states. We denote by $\boldsymbol{\theta}=(\boldsymbol{\theta}_1,...,\boldsymbol{\theta}_N)$ the set of parameters specifying the set of linear models indexed by the different states.

For convenience, we use the notation $\lambda = (\mathbf{A},\boldsymbol{\pi},\boldsymbol{\delta},\boldsymbol{\theta})$ to specify an stochastic process called {\bf Piecewise Linear Hidden Markov Model} (PLHMM). A PLHMM $\lambda$ is a point in a parameter manifold $\Lambda$. PLHMM is a sort of Hidden semi-Markov Model, in the sense that a transition from the state $S_i$ depends on the amount of time elapsed since entering into $S_i$.

Let us consider PLHMM from a generative point of view by ancestral sampling. We first choose an initial state $q_1=S_i$ according to the initial state distribution $\boldsymbol{\pi}$. Now we choose a duration $d_1$ according to the state duration density $p_i(d_1)$. We then choose a set of observations $v_1,v_2,...,v_{d_1}$ according to the joint observable probability distribution $b_i(v_1,v_2,...,v_{d_1})$. We also assume that these observations are drawn independently from the distribution (\ref{eq:emission}), so that
 \begin{displaymath}
 b_i(v_1,v_2,...,v_{d_1})=\prod ^{d_1} _{t=1} \mathcal{N}(v_t|y_i(t),\beta_i^{-1}).
\end{displaymath}
Then we choose the next state, $q_2=S_j$ according to the state transition probability $a_{ij}$ and so on. We formally specify a PLHMM in the following definition.

\begin{definition}
 The pair of sequences of random variables $q_{1:S}=(q_s)_{s=1}^S$ and $v_{1:T}=(v_t)_{t=1}^T$ are distributed according to a Piecewise Linear Hidden Markov Model, written $(q_{1:S},v_{1:T})\sim \text{PLHMM} (\lambda)$, when they follow the generative process
 \begin{align*}
  q_1 & \sim \pi, \\
  q_{s+1}|q_{s} & \sim  a_{q_{s}q_{s+1}} &  &s=1,...,S-1  \\
  d_s| q_s & \sim  p_{q_s}(d_s) &  &s=1,...,S-1  \\
  v_t| q_s & \stackrel{iid}{\sim}  \mathcal{N}(v_t|\boldsymbol{w}_{q_s}^T\boldsymbol{\phi}(t),\beta_{q_s}^{-1}) &  &\sum ^{s-1} _{r=0} d_r < t \leq \sum ^s _{r=1} d_r.
 \end{align*}
\end{definition}

Given a sequence $v_{1:T}$, we can calculate the likelihood $P(v_{1:T}|\lambda)$, given the model $\lambda$, so that 
\begin{eqnarray}
P(v_{1:T}|\lambda) & =& \sum _{q_1,...,q_S} \sum _{d_1,...,d_S} \pi_{q_1} p_{q_1}(d_1) \prod ^{d_1} _{t=1} b_{q_1}(v_t) \cdots \nonumber \\
 & & \cdot ~a_{q_1q_2} p_{q_2}(d_2) \prod ^{d_2} _{t=d_1+1} b_{q_2} (v_t) \cdots \nonumber \\ 
 & & \cdot ~a_{q_{S-1}q_{S}} p_{q_S}(d_S) \!\! \prod ^{T} _{t=d_1+...+d_{S-1}+1} \!\! b_{q_S} (v_t),
 \label{eq:likelihood_raw}
\end{eqnarray}
where it is assumed that $\sum_{s=0}^S d_s=T$.

\section{Efficient computation of the likelihood}

We can efficiently compute this likelihood function by computing the joint probability $P(v_{1:T},q_s=S_j|\lambda)$ at any arbitrary state index $s$, and summing over all state sequences:
\begin{equation}
 P(v_{1:T}|\lambda)=\sum ^N _{j=1} P(v_{1:T},q_s=S_j|\lambda)
\label{eq:likelihood}
 \end{equation}
The term satisfies $P(v_{1:T},q_s=S_j|\lambda)=P(v_{1:T}|q_s = S_j,\lambda)P(q_s = S_j|\lambda)$. Following \cite{Rabiner89} let us suppose that $q_s$ ends at $t$, so $t=\sum_{r=1} ^{s} d_r$. Then, the conditional likelihood of $v_1,...,v_t$ is independent of $v_{t+1},...,v_T$, given $q_s=S_j$, in the application of d-separation, and (\ref{eq:likelihood}) can be rewritten as
\begin{eqnarray*}
 P(v_{1:T}|\lambda) & = & \sum ^N _{j=1} P(v_1,...,v_t|q_s=S_j,\lambda) \nonumber \\
 & & \hspace{10mm} P(v_{t+1},...,v_{T}|q_s=S_j,\lambda)P(q_s=S_j|\lambda) \nonumber \\
 & = & \sum ^N _{j=1} P(v_1,...,v_t,q_s=S_j|\lambda) \nonumber \\
 & & \hspace{30mm} P(v_{t+1},...,v_{T}|q_s=S_j,\lambda) \nonumber \\
 & = & \sum ^N _{j=1} \alpha_{t}(j) \beta_{t}(j)\hspace{15 mm} t\in[1,T]
\end{eqnarray*}
where
\begin{displaymath}
 \alpha_t(j) \triangleq P(v_1,...,v_t,q_s=S_j|\lambda)
\end{displaymath}
and 
\begin{displaymath}
 \beta_t(j) \triangleq P(v_{t+1},...,v_{T}|q_s=S_j,\lambda).
\end{displaymath}
Furthermore,
\begin{displaymath}
 \alpha_t(j) = \sum ^N _{\substack{i=1 \\ i\neq j}} P(v_1,...,v_t,q_{s-1}=S_i,q_s=S_j|\lambda)
\end{displaymath}
and
\begin{eqnarray*}
 P(v_1,...,v_t,q_{s-1}=S_i,q_s=S_j|\lambda) & = & \nonumber \\
 & & \hspace{-35mm} = \sum _{d_s<t} P(v_1,...,v_t,q_{s-1}=S_i,q_s=S_j,d_s|\lambda),
\end{eqnarray*}
with
\begin{eqnarray*}
P(v_1,...,v_t,q_{s-1}=S_i,q_s=S_j,d_s|\lambda) & = & \nonumber \\
& & \hspace{-28mm} = \alpha_{t-d_s}(i)a_{ij}p_j(d_s)\prod ^t _{r=t-d_s+1} b_j(v_r),
\end{eqnarray*}
where we sum over all the possible duration values $d_s$ of the state $S_j$. Since the initial state may last for $d_1$ observations prior to the occurrence of any state transitions, previous equation should be initialized, resulting in
\begin{eqnarray}
 \alpha_t(j) & = & \pi_jp_j(t)\prod ^{t} _{r=1}b_j(v_r) + \nonumber \\
 &  & \hspace{5mm} + \sum ^N _{\substack{i=1 \\ i\neq j}} \sum _{d<t} \alpha_{t-d}(i) a_{ij} p_j(d) \prod ^t _{r=t-d+1} b_j(v_r),
 \label{eq:alpha}
\end{eqnarray}
where the index $t$ in the first term plays the role of duration for the first state \cite{Rabiner89}. Similarly,
\begin{equation}
 \beta_t(i) = \sum ^N _{\substack{j=1 \\ j\neq i}} \sum _{d\leq T-t}a_{ij}p_j(d)\beta_{t+d}(j)\prod ^{t+d} _{r=t+1} b_j(v_r),
\label{eq:beta}
\end{equation}
with $\beta_T(i)\triangleq 1$, $1\leq i\leq N$, meaning that any future evolution of the time series after $t=T$ is fully probable. The likelihood function can be written as
\begin{eqnarray}
 P(v_{1:T}|\lambda) & = &\sum_{j=1}^N \pi_jp_j(t)\prod ^t _{r=1} b_j(v_r) \beta _t(j) + \nonumber \\
 & & \hspace{-13mm} + \sum ^N _{\substack{j=1}} \sum ^N _{\substack{i=1 \\ i\neq j}} \sum _{d<t}\alpha_{t-d}(i)a_{ij}p_j(d)\!\!\prod ^t _{r=t-d+1} \!b_j(v_r) \beta _t(j).
 \label{eq:likelihood_alpha_beta}
\end{eqnarray}
We can also write the likelihood function as
\begin{equation}
 P(v_{1:T}|\lambda)=\sum ^N _{j=1} \alpha_T(j).
\end{equation}
Equations (\ref{eq:alpha}) and (\ref{eq:beta}) are an adaptation from Baum and colleagues \cite{Baum67} and define what is known as the forward-backward procedure.

\section{Maximum likelihood estimation of the model}
Having observed a sequence $v_{1:T}$, the training problem is that of determining the model $\lambda = (\mathbf{A},\boldsymbol{\pi},\boldsymbol{\delta},\boldsymbol{\theta})$ that best fits this sequence. Let us assume that the number of states $S$ for the model is fixed a priori, enforcing a subsequent constraint $\sum_{s=1} ^S d_s=T$ for the corresponding state sequence $q_{1:S}$. The likelihood function $P(v_{1:T}|\lambda)$ allows us to determine the parameters of the model using maximum likelihood.

Unfortunately, there is no analytical solution to likelihood maximization. We can, however, choose $\lambda$ such that $P(v_{1:T}|\lambda)$ is locally maximized, using the Baum-Welch iterative procedure \cite{Baum70}, or using gradient techniques \cite{Levinson83}, guaranteeing monotonic increase in the likelihood as the procedure iterates.

The Baum-Welch iterative procedure is a reestimation procedure, that starts with an initial guess of the model $\lambda$ and provides reestimation formulas that lead to increase $P(v_{1:T}|\lambda)$ except if we are at a critical point of $P(v_{1:T}|\lambda)$. Baum-Welch procedure is based on the following lemma:

\begin{lemma}
 \cite{Baum72} Let $u_n$, $1\leq n\leq W$ be positive real numbers, and let $v_n$, $1\leq n\leq W$ be nonnegative real numbers such that $\sum_n v_n>0$. Then from concavity of the log function it follows that
\end{lemma}
 \begin{displaymath}
  \text{ln}~ \left(\frac{\sum _n v_n}{\sum _n u_n}\right)\geq \frac{1}{\sum _n u_n}\left[\sum _n(u_n \text{ln}~ v_n - u_n \text{ln}~ u_n)\right]
 \end{displaymath}
Let $W$ be the number of state sequences $q_{1:S}$ of length $S$. For the $nth$ sequence $q^n_{1:S}=(q_1=S_{n_1},q_2=S_{n_2},...,q_S=S_{n_S})$ let $u_n$ be the joint probability $u_n=P(v_{1:T},q^n_{1:S}|\lambda)$. Let $v_n$ be the joint probability $v_n=P(v_{1:T},q^n_{1:S}|\bar{\lambda})$ conditioned on a different model $\bar{\lambda}$. Then 
\begin{eqnarray*}
 \sum_{n=1} ^S u_n & = & P(v_{1:T}|\lambda) \\
 \sum_{n=1} ^S v_n & = & P(v_{1:T}|\bar{\lambda}) 
\end{eqnarray*}
According to the above lemma
\begin{displaymath}
 \text{ln}~ \frac{P(v_{1:T}|\bar{\lambda})}{P(v_{1:T}|\lambda)} \geq \frac{1}{P(v_{1:T}|\lambda)} \big[Q(\lambda,\bar{\lambda})-Q(\lambda,\lambda)\big]
\end{displaymath}
where
\begin{equation}
 Q(\lambda,\bar{\lambda})\triangleq \sum _n u_n \text{ln}~ v_n  = \sum_{n=1} ^S P(v_{1:T},q^n_{1:S}|\lambda)~ \text{ln}~ P(v_{1:T},q^n_{1:S}|\bar{\lambda})
\label{eq:q}
 \end{equation}
As a consequence, $Q(\lambda,\bar{\lambda})\geq Q(\lambda,\lambda)$ implies that $P(v_{1:T}|\bar{\lambda})\geq  P(v_{1:T}|\lambda)$, with equality iff $\lambda$ is a critical point. The term in (\ref{eq:likelihood_raw}) is, in fact, the joint density $P(v_{1:T},q^n_{1:S}|\lambda)$ for a particular assignment to the sequence of states and corresponding duration. Reordering terms in this term we can obtain
\begin{eqnarray*}
 P(v_{1:T},q^n_{1:S}|\bar{\lambda}) & = & \bar{\pi}_{n_1} \prod ^{S-1} _{s=1} \bar{a}_{n_sn_{s+1}} \prod ^S _{s=1} \bar{p}_{n_s}(d_s) \prod ^{d_1} _{t=1} b_{n_1}(v_t) \cdot \nonumber \\
 & & \hspace{2mm} \cdot \prod ^{d_2} _{t=d_1+1} b_{n_2} (v_t)~ \dots  \hspace{-5mm} \prod ^{T} _{t=d_1+...+d_{S-1}+1} b_{n_S} (v_t)
\end{eqnarray*}
Thus,
\begin{eqnarray*}
 \text{ln}~P(v_{1:T},q^n_{1:S}|\bar{\lambda}) & = & \text{ln}~\bar{\pi}_{n_1} + \sum ^{S-1} _{s=1} \text{ln}~\bar{a}_{n_sn_{s+1}} + \sum ^S _{s=1} \text{ln}~\bar{p}_{n_s}(d_s) \nonumber \\
 &  & \hspace{5mm} + \sum ^{d_1} _{t=1} \text{ln}~b_{n_1}(v_t) + \sum ^{d_1+d_2} _{t=d_1+1} \text{ln}~b_{n_2} (v_t) + \nonumber \\
 &  & \hspace{10mm} + \cdots + \sum ^{T} _{t=d_1+ ...+d_{S-1}+1} \text{ln}~b_{n_S} (v_t)
\end{eqnarray*}
Substituting this in (\ref{eq:q}) and regrouping terms it can be seen
\begin{eqnarray*}
Q(\lambda,\bar{\lambda}) & = & \sum _{i=1} ^N \gamma _{0i}~ \text{ln}~ \bar{\pi}_{i} + \sum _{i=1} ^N \sum _{j=1} ^N \gamma _{ij}~ \text{ln}~ \bar{a}_{ij} + \nonumber \\
 &  & \hspace{15mm} + \sum ^N _{i=1} \sum _{d} \eta_{id}~ \text{ln}~ \bar{p}_i(d)+ \sum _{s=1} ^S Q_{b_{n_s}}(\lambda,\bar{\lambda})
\end{eqnarray*}
Next we explain the different terms of this expression. For the first term
\begin{displaymath}
 \gamma_{0i} = \sum _{n=1} ^W  P(v_{1:T},q^n_{1:S}|\lambda)~\delta(q_1=S_i),
\end{displaymath}
where we sum over all the state sequences beginning with the assignment $q_1=S_i$. In the above expression $\delta(.)$ is the Kronecker delta function. Thus, $\gamma _{0i}$ can be interpreted as the expected number of sequences beginning with state $S_i$ since the expectation of a binary random variable is defined as the probability of taking the value 1.

For the second term
\begin{displaymath}
 \gamma _{ij} = \sum _{n=1} ^W \sum _{s=1} ^{S-1} P(v_{1:T},q^n_{1:S}|\lambda)~\delta(q_s=S_{i},q_{s+1}=S_{j}),
\end{displaymath}
where the inner term is the expected number of transitions from state $S_i$ to state $S_j$ in a given state sequence, and the term $\gamma_{ij}$ can be interpreted as the expected number of transitions from state $S_i$ to state $S_j$ in all the different state sequences.

For the third term we firstly assume a discrete state duration distribution, and hence
\begin{displaymath}
 \eta_{id} = \sum _{n=1} ^W \sum _{s=1} ^{S} P(v_{1:T},q^n_{1:S}|\lambda)~\delta(q_s=S_i,d_s=d),
\end{displaymath}
being $\eta_{id}$ the expected number of times the state $S_i$ occurs with duration $d$ in all the different state sequences. The following section gives an alternative with a gamma distribution.

Maximization with respect to $\boldsymbol{\pi}$, $\boldsymbol{A}$ and $\boldsymbol{\delta}$ can be achieved by using appropriate Lagrange multipliers, subject to the constraints $\sum_i \pi_i=1$, $\sum_j a_{ij}=1$ and $\sum_d p_i(d)=1$, respectively. Applying the Lagrange method to $\boldsymbol{\pi}$ we obtain
\begin{displaymath}
 \frac{\partial}{\partial \overline{\pi}_i} \bigg[ Q(\lambda,\bar{\lambda}) -\mu \bigg(\sum_i \overline{\pi}_i-1\bigg)\bigg] = \frac{\gamma_{0i}}{\overline{\pi}_i}-\mu = 0,
\end{displaymath}
Multiplying by $\bar{\pi}_i$ and summing over $i$ gives $\mu = \sum_i \gamma_{0i}$, hence
\begin{displaymath}
 \overline{\pi}_i= \frac{\gamma_{0i}}{\sum \limits _i \gamma_{0i}}
\end{displaymath}
We can express this reestimate in terms of the forward and backward probabilities. Using the definition of $\gamma_{0i}$, we can rewritten the likelihood $P(v_{1:T}|\lambda)$ given by (\ref{eq:likelihood_raw}) in terms of the definition of $\beta$ in (\ref{eq:beta}), and we have
\begin{eqnarray*}
 \gamma_{0i} & = & \sum _{n=1} ^W P(v_{1:T},q^n_{1:S}|\lambda)~\delta(q_1=S_i) = \nonumber \\
 &  & \hspace{25mm} = \sum _{d_1<t} \pi_i p_i(d_1) \prod ^{d_1} _{t=1} b_i(v_t) \beta_{d_1}(i).
\end{eqnarray*}
Summing over all the possible states
\begin{displaymath}
 \sum_i \gamma_{0i} = \sum_{i=1} ^N \alpha_{d_1}(i)\beta_{d_1}(i)= P(v_{1:T}|\lambda).
\end{displaymath}
Then
\begin{equation}
\overline{\pi}_i= \frac{1}{P(v_{1:T}|\lambda)} \sum \limits_{d_1<t} \pi_i p_i(d_1) \prod \limits^{d_1} _{s=1} b_i(v_s) \beta_{d_1}(i)
\label{eq:pi_reestimate}
\end{equation}
Applying the Lagrange method to $\boldsymbol{A}$ we obtain
\begin{displaymath}
 \overline{a}_{ij}=\frac{\gamma _{ij}}{\sum \limits_j \gamma_{ij}}
 \end{displaymath}
We can also express this reestimate in terms of the forward and backward probabilities. Using the definition of $\gamma_{ij}$, we can rewritten the likelihood $P(\boldsymbol{v}|\lambda)$ given by (\ref{eq:likelihood_raw}) in terms of the definition of $\alpha$ and $\beta$, and we have
\begin{eqnarray*}
 \gamma _{ij} & = & \sum _{n=1} ^S \sum _{r=1} ^{M-1} P(\boldsymbol{q}_n,\boldsymbol{v}|\lambda)~\delta(q_r=S_{i},q_{r+1}=S_{j}) \\
 & = & \sum_{t=1} ^T \sum _{d_r<t} \alpha_{t-d_r}(i)a_{ij}p_j(d_r)\prod ^t _{s=t-d_r+1} b_j(v_s) \beta _t(j)
\end{eqnarray*}
Summing over all the possible states
\begin{displaymath}
 \sum_{j=1} ^N \gamma_{ij} = \sum_{t=1} ^T \alpha_t(i) \beta_t(i).
\end{displaymath}
Then
\begin{equation}
 \overline{a}_{ij}= \frac{\sum \limits_{t=1} ^T \sum \limits_{d_r<t} \alpha_{t-d_r}(i)a_{ij}p_j(d_r)\prod \limits^t _{s=t-d_r+1} b_j(v_s) \beta _t(j)}{\sum \limits_{t=1} ^T \alpha_t(i) \beta_t(i)}
\end{equation}
This reestimation formula can be interpreted as the fraction between the expected number of transitions from state $S_i$ to state $S_j$ and the expected number of transitions from state $S_i$, along the sequence $\boldsymbol{v}$.

Applying the Lagrange method to $\boldsymbol{d}$ we obtain
\begin{displaymath}
 \overline{p}_{j}(d)=\frac{\eta _{jd}}{\sum \limits_d \eta_{jd}}
 \end{displaymath}
We can also express this reestimate in terms of the forward and backward probabilities. Thus
\begin{eqnarray*}
 \eta_{jd} & = & \sum _{n=1} ^S \sum _{r=1} ^{M} P(\boldsymbol{q}_n,\boldsymbol{v}|\lambda)~\delta(q_r=S_j,d_r=d)\\
 & = & \sum_{t=1} ^T \sum_{i=1} ^N \alpha_{t-d} (i) a_{ij} p_j(d) \prod _{s=t-d+1} ^t b_j(v_s) \beta_t (j).
\end{eqnarray*}
 Summing over all the possible duration values we obtain
 \begin{equation}
  \overline{p}_j(d)=\frac{\sum \limits_{t=1} ^T \sum \limits_{i=1} ^N \alpha_{t-d} (i) a_{ij} p_j(d) \prod \limits _{s=t-d+1} ^t b_j(v_s) \beta_t (j)}{\sum \limits_{t=1} ^T \alpha_{t} (j) \beta_t (j)},
 \end{equation}
under the constraint $d<t$. This reestimation formula can be interpreted as the fraction between the expected number of times the state $S_j$ occurs with duration $d$ and the expected number of times the state $S_j$ occurs with any duration.
 
For the fourth term
\begin{displaymath}
Q_{b_{n_r}}(\lambda,\bar{\lambda}) = \sum _{i=1} ^N \sum ^{d_1+...+d_r} _{s=d_1+...+d_{r-1}+1} \xi_{ri}~  \text{ln}~b_{i} (v_s),
\end{displaymath}

where
\begin{displaymath}
 \xi_{ri} = \sum _{n=1} ^S  P(\boldsymbol{q}_n,\boldsymbol{v}|\lambda)~\delta(q_r=S_i),
\end{displaymath}
being $\xi_{ri}$ the expected number of sequences such that the r-th state is $S_i$. Note that $\xi_{1i}=\gamma_{1i}$. According to equation (\ref{eq:emission}), and denoting $\tilde{d}_r=d_1+...+d_r$ to keep the notation uncluttered, we have
\begin{eqnarray*}
 Q_{b_{n_r}}(\lambda,\bar{\lambda}) & = & \sum _{i=1} ^N \xi_{ri} \sum ^{\tilde{d}_r} _{s=\tilde{d}_{r-1}+1}  \text{ln}~ \mathcal{N}(v_s|\boldsymbol{w}_i^T\boldsymbol{\phi}(s),\beta_i^{-1}) \\
 & = & \sum _{i=1} ^N \xi_{ri} \bigg( \frac{d_r}{2}~ \text{ln}~ \beta - \frac{d_r}{2}~ \text{ln}~(2\pi) - \nonumber \\
 &  & \hspace{20mm} - \frac{\beta}{2} \sum ^{\tilde{d}_r} _{s=\tilde{d}_{r-1}+1} \big(v_s- \boldsymbol{w}_i^T\boldsymbol{\phi}(s)\big)^2\bigg).
\end{eqnarray*}
Maximization with respect to $\boldsymbol{w}_i$ is equivalent to minimizing the classical sum-of-squares error function of linear regression analysis given by the third term in the above expression. The result of applying the gradient is
\begin{displaymath}
 \nabla _i  Q_{b_{n_r}}(\lambda,\bar{\lambda}) = \xi_{ri}~ \beta \sum ^{\tilde{d}_r} _{s=\tilde{d}_{r-1}+1} (v_s- \boldsymbol{w}_i^T\boldsymbol{\phi}(s)) \boldsymbol{\phi}(s)^T.
\end{displaymath}
Setting this gradient to zero results in
\begin{displaymath}
\sum ^{\tilde{d}_r} _{s=\tilde{d}_{r-1}+1} \!\!\! v_s \boldsymbol{\phi}(s)^T - \boldsymbol{w}_i^T \!\!\! \sum ^{\tilde{d}_r} _{s=\tilde{d}_{r-1}+1}\!\!\! \boldsymbol{\phi}(s) \boldsymbol{\phi}(s)^T =0,
\end{displaymath}
and solving for $\boldsymbol{w}_i$ we obtain
\begin{displaymath}
 \overline{\boldsymbol{w}}_i = (\boldsymbol{\Phi} ^T \boldsymbol{\Phi})^{-1} ~\boldsymbol{\Phi}
 ^T \boldsymbol{v}_r
\end{displaymath}
where $\boldsymbol{v}_r$ is the fragment of the time series $\boldsymbol{v}$ from $\tilde{d}_{r-1}+1$ to $\tilde{d}_{r}$. Symbol $\boldsymbol{\Phi}$ represents the following matrix
\begin{displaymath}
 \boldsymbol{\Phi} = \left( \begin{array}{cccc}
                \phi_0(v_{\tilde{d}_{r-1}+1}) & \phi_1(v_{\tilde{d}_{r-1}+1}) & \cdots & \phi_M(v_{\tilde{d}_{r-1}+1})\\
                \phi_0(v_{\tilde{d}_{r-1}+2}) & \phi_1(v_{\tilde{d}_{r-1}+2}) & \cdots & \phi_M(v_{\tilde{d}_{r-1}+2})\\
                \vdots & \vdots & \ddots & \vdots \\
                \phi_0(v_{\tilde{d}_{r}}) & \phi_1(v_{\tilde{d}_{r}}) & \cdots & \phi_M(v_{\tilde{d}_{r}})\\
               \end{array} \right)
\end{displaymath}

Maximization with respect to $\beta$ gives
\begin{displaymath}
 \frac{1}{\overline{\beta}} = \frac{1}{d_r} \sum ^{\tilde{d}_r} _{s=\tilde{d}_{r-1}+1} \big(v_s- \overline{\boldsymbol{w}}_i^T\boldsymbol{\phi}(s)\big)^2,
\end{displaymath}
that is, the inverse of the precision is reestimated as the residual variance of the time series values around the regression function, as it was expected.

These reestimation formulas can be used for training a PLHMM in an Expectation-Maximization fashion, as an extension of the traditional Baum-Welch algorithm for training Hidden Markov Models. 

\subsection{State duration estimation through Gamma distribution}
An important drawback to using the above representation of state duration density is its inherent complexity, as the number of parameters associated with the duration of each state is the cardinal of its domain, i.e., the number of samples of the time series.

An alternative is to use a parametric family of continuous probability density functions to obtain the duration probabilities. The Gamma distribution is well suited to this end, and indeed, it is frequently used in science and technology to model waiting times:
\begin{displaymath}
 p(d) = \frac{\eta^{\nu}}{\Gamma(\nu)}d^{\nu-1}e^{-\eta d},
\end{displaymath}
with a \textit{shape parameter} $\nu$ and a \textit{rate parameter} $\eta$. The mean value of the distribution is $\nu/\eta$ and the variance is $\nu/\eta ^2$.

In order to apply the Gamma distribution to a discrete problem we compute the integral from $d$ to $d+1$, by using the  lower incomplete Gamma function. The resulting distribution satisfies normalization:
\begin{displaymath}
\sum_{d=1}^{T}p_j(d)=1,\qquad 1\leq j \leq N.
\end{displaymath}

Reestimation formulas should be obtained for $\nu$ and $\eta$ so as to train the model. Following \cite{Levinson86}, reestimation is carried out by maximizing likelihood in equation (\ref{eq:likelihood_alpha_beta}). Straightforward differentiation of Gamma distribution for each state yields 
\begin{displaymath}
 \frac{\partial p_j(d) }{\partial \eta_j} = \frac{\eta_j^{\nu_j}}{\Gamma(\nu_j)}d^{\nu_j-1}e^{-\eta_j d}\left[\frac{\nu_j}{\eta_j}- d\right] = p_j(d)\left[\frac{\nu_j}{\eta_j}- d\right]
\end{displaymath}
for the rate parameter $\eta$, and 
\begin{eqnarray*} \frac{\partial p_j(d) }{\partial \nu_j} & = & \frac{\eta_j^{\nu_j}}{\Gamma(\nu_j)}d^{\nu_j-1}e^{-\eta_j d}\left[ \log(\eta_j d)-\frac{\Gamma'(\nu_j)}{\Gamma(\nu_j)}\right] \nonumber \\ 
 & & \hspace{30mm} = p_j(d)\left[\log(\eta_j d)-\psi(\nu_j) \right]
\end{eqnarray*}
for the shape parameter $\nu$. We maximize the likelihood with respect to $\eta$, obtaining
\begin{eqnarray*}
 \frac{\partial P(v_{1:T}|\lambda)}{\partial \eta_j} & = & \pi_jp_j(t)\left[\frac{\nu_j}{\eta_j}- t\right]\prod ^t _{r=1} b_j(v_r) \beta _t(j) + \nonumber \\
& &  \hspace{8mm} + \sum ^N _{\substack{i=1 \\ i\neq j}} \sum _{d<t}\alpha_{t-d}(i)a_{ij}p_j(d)\left[\frac{\nu_j}{\eta_j}- d\right] \cdot \nonumber \\
& & \hspace{25mm} \cdot \prod ^t _{r=t-d+1} b_j(v_r) \beta _t(j) = 0.
\end{eqnarray*}

Multiplying by $\eta_j$ and summing along $t$ we obtain
\begin{displaymath}
\overline{\eta_j} = \frac{\nu_j \sum \limits_{t=1} ^T \alpha_t(j)\beta_t(j)}{\sum \limits_{t=1}^T \xi_t},
\end{displaymath}

where
\begin{eqnarray*}
\xi_t & = & \pi_j p_j(t) t \prod \limits^t _{r=1} b_j(v_r) \beta _t(j) + \sum \limits_{d<t} d \sum \limits_{\substack{i=1 \\ i\neq j}}^N \alpha_{t-d}(i)a_{ij}p_j(d) \cdot \nonumber \\
& & \hspace{47mm} \cdot \prod \limits^t _{r=t-d+1} b_j(v_r) \beta _t(j).
\end{eqnarray*}

Similarly, we maximize the likelihood with respect to $\nu_j$, obtaining
\begin{eqnarray*}
\frac{\partial P(v_{1:T}|\lambda)}{\partial \nu_j} & = & \pi_jp_j(t)\left[log(t\eta_j)-\psi(\nu_j)\right]\prod ^t _{r=1} b_j(v_r) \beta _t(j) + \nonumber \\
&  & + \sum ^N _{\substack{i=1 \\ i\neq j}} \sum _{d<t}\alpha_{t-d}(i)a_{ij}p_j(d)\left[log(t\eta_j)-\psi(\nu_j)\right] \cdot \nonumber \\ 
&  & \hspace{30mm} \cdot \prod ^t _{r=t-d+1} b_j(v_r) \beta _t(j) =0.
\end{eqnarray*}

Summing along $t$ we obtain
\begin{equation}
\psi(\overline{\nu_j}) = \frac{\sum \limits_{t=1}^T \zeta_t}{\sum \limits_{t=1} ^T \alpha_t(j)\beta_t(j)},
\label{eq:digamma}
\end{equation}

where 
\begin{eqnarray*}
\zeta_t & = & \pi_j p_j(t) \log(t\eta_j) \prod \limits^t _{r=1} b_j(v_r) \beta _t(j) + \nonumber \\
&  & \sum \limits_{d<t} \log(d\eta_j) \sum \limits_{\substack{i=1 \\ i\neq j}}^N \alpha_{t-d}(i)a_{ij}p_j(d)\prod \limits^t _{r=t-d+1} b_j(v_r) \beta _t(j),
\end{eqnarray*}
and $\psi$ is the well known digamma function. Equation (\ref{eq:digamma}) can be solved numerically by Newton-Raphson method, starting with an initial guess $y$, such that
\begin{eqnarray*}
y = \left\lbrace \begin{array}{cl}
     {\displaystyle e^{x}+\frac{1}{2}}& \text{ if } x\geq -2.22  \\ 
     {\displaystyle \frac{-1}{x+\psi^{(0)}(1)}}& \text{ otherwise},
\end{array}\right.
\end{eqnarray*}

where $\psi(\overline{\nu}_j)=x$. From that initial value we iterate according to
\begin{displaymath}
y^* = y-\frac{\psi^{(0)}(y)-x}{\psi^{(1)}(y)}.
\end{displaymath}

\begin{figure*}[htb]
\centering
\subfloat[Normal beat, recording 100]{\includegraphics[width=1.5in]{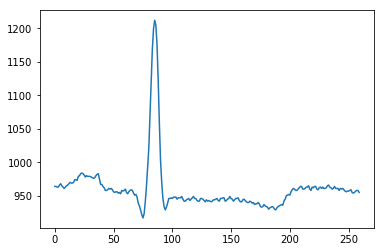}%
\label{fig_first_beat}}
\hfil
\subfloat[Normal beat, rec. 103]{\includegraphics[width=1.5in]{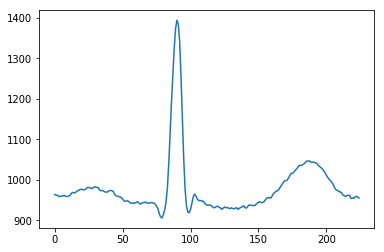}%
\label{fig_second_beat}}
\hfil
\subfloat[Ventricular beat, rec. 119]{\includegraphics[width=1.5in]{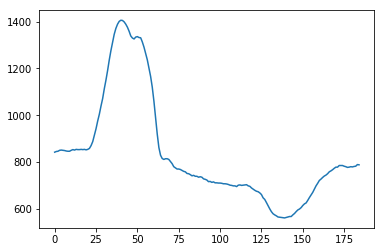}%
\label{fig_second_beat}}
\hfil
\subfloat[Normal beat, rec. 106]{\includegraphics[width=1.5in]{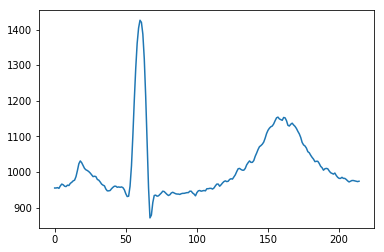}%
\label{fig_second_beat}}
\caption{Different heartbeat morphologies from MIT-BIH Arrhythmia Database.}
\label{fig:heartbeat_morphologies}
\end{figure*}

\section{Experimental results}
In this section, we apply PLHMM to different time series from electrocardiographic monitoring with the aim of illustrating its viability in a real setting. Experiments were performed using the MIT-BIH Arrhythmia Database \cite{Moody01} from the Physionet initiative \cite{Goldberger00}. This database can be considered the gold standard for arrhythmia and heartbeat classification, and it has been used in most of the published research on machine learning. Three different features were tested: i) {\em abstraction}, as the ability of PLHMM to model a heartbeat morphology from one single example; ii) {\em recognition}, as the ability of the model to correctly identify different morphologies; and iii) {\em performance}, as the computational time it requires to train the model by using each state duration representation.

Along the experiments a left-to-right topology was used (for any state, once left, cannot be later revisited). We fixed the number of states at N=7. A common Hermite orthonormal basis was used for modelling every observable probability distribution, due to their resemblance to the morphological constituents of a heartbeat. We fixed the number of Hermite coefficients for each state at $(3,5,1,6,1,5,3)$ in order to test a range of different expressiveness levels.

\subsection{Abstraction tests}
\begin{figure}[!t]
\centering
\subfloat[]{\includegraphics[width=2in]{figures/106_Original.png}%
\label{fig_first_case}}
\hfil
\subfloat[]{\includegraphics[width=2in]{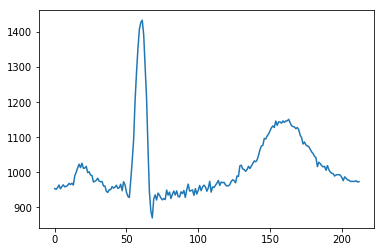}%
\label{fig_second_case}}
\hfil
\subfloat[]{\includegraphics[width=2in]{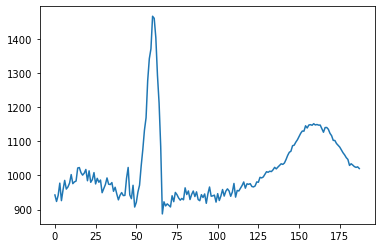}%
\label{fig_third_case}}
\caption{Simulation results for the model. (a) Original heart beat [Source: MIT-BIH Arrhythmia Database, recording: 106, lead: MLII]; (b) Time series sampled from the model, where duration is represented by a discrete probability distribution; (c) Time series sampled from the model, where duration is represented by the Gamma distribution.}
\label{fig:abstraction}
\end{figure}

A set of exploratory experiments was performed by using PLHMM as a generative model. Figure \ref{fig:abstraction} shows an original normal heartbeat, and the simulation results for the model, testing the two different ways of representing duration probability distributions. It should be noted the presence of Gaussian noise in simulation results, trying to mimic some high-frequency processes in the original signal with a possible physiological interpretation which is out of the scope of the present paper. Similar results were obtained on different beat morphologies from ECG recordings 100, 103 and 119 (see Figure \ref{fig:heartbeat_morphologies}).

%Para comenzar el ajuste utilizamos un conjunto de parámetros iniciales estimados a partir de una segmentación conocida del latido. Aplicando este conocimiento experto generamos unas distribuciones iniciales que nos permiten agilizar la convergencia. La forma de generar estos parámetros iniciales de la función gamma usando la segmentación se basa en el sistema de ecuaciones:

%\begin{eqnarray*}
%(\eta_j,\nu_j) = \left\lbrace \begin{array}{l}
%     {\displaystyle mode = \frac{\nu_j-1}{\eta_j}} \\ 
%     {\displaystyle var = \frac{\nu_j}{\eta_j^2}}
%\end{array}\right.
%\end{eqnarray*}

\subsection{Recognition tests}
\begin{figure*}[!htb]
\centering
\includegraphics[width=\textwidth]{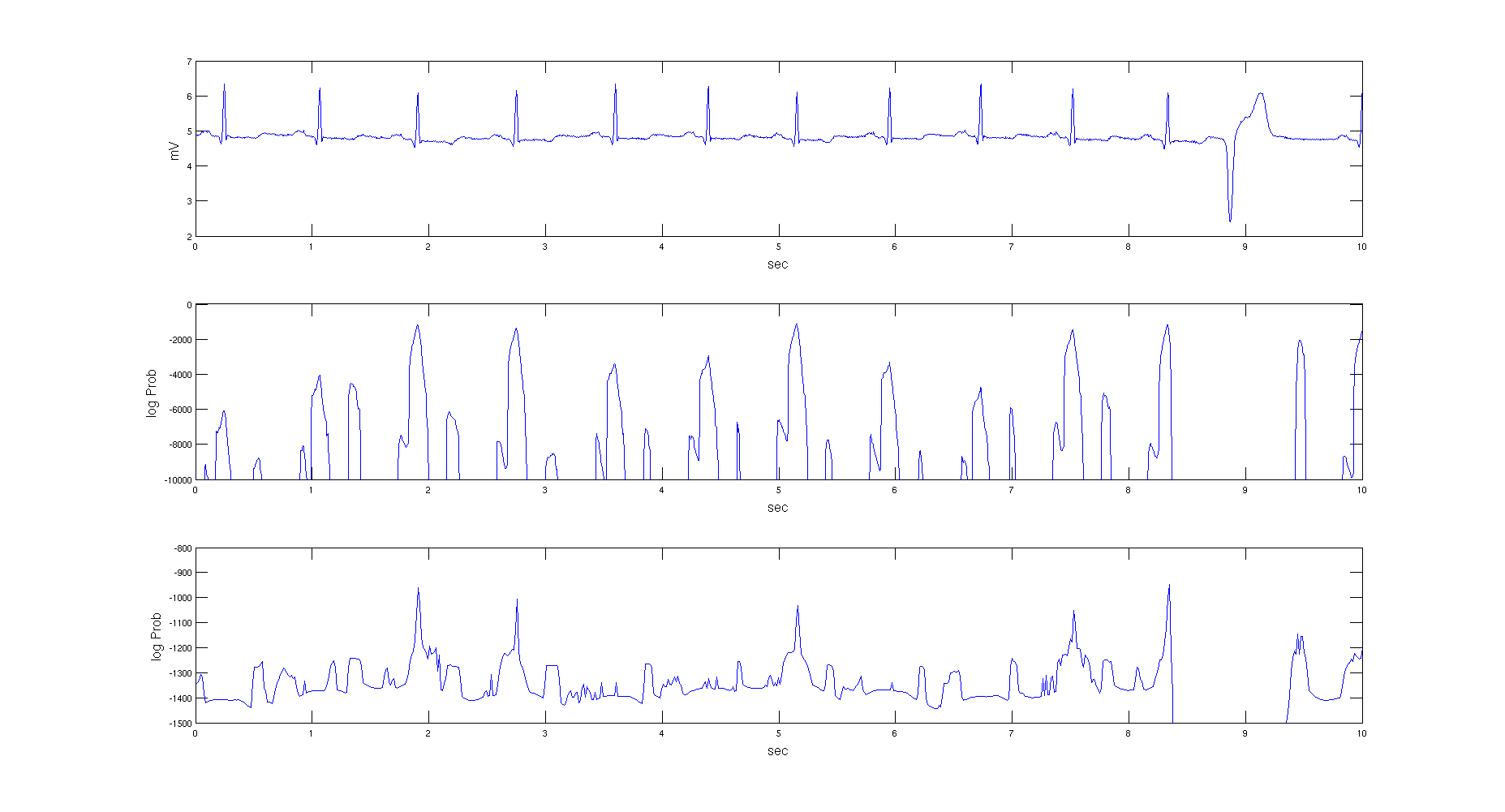}
\caption{Results of the detection of a normal heartbeat. The upper panel shows an ECG strip with a predominant normal rhythm and one ventricular beat at the end [Source: MIT-BIH Arrhythmia Database, recording: 100, lead: MLII, between 25:10.000 and 25:20.000]. The middle panel shows the result of the likelihood computed for a PLHMM where each duration is represented by a discrete probability distribution. In the bottom panel each duration is represented by a Gamma distribution.}
\label{fig:recognition}
\end{figure*}

A set of exploratory experiments was performed by using PLHMM as a recognition tool. Figure \ref{fig:recognition} shows a short ECG strip with predominant normal rhythm and a ventricular beat in 12th position. After training the model with the 2nd beat in the strip (a normal beat), recognition is performed by computing likelihood along a sliding window of constant width (width = 260 samples). Experiments show a significant likelihood for each normal beat, and a negligible likelihood for the ventricular beat. It should be noted that three local maximum values are identified for each normal beat, resulting from the alignment of the model with the three outstanding constituents of the heartbeat: the P wave, the QRS complex and the T wave. The largest of them is aligned with the QRS complex, the most prominent wave in the cardiac cycle. After the ventricular beat a local maximum is identified, corresponding to the next beat. These results can be easily improved by enforcing alignment with QRS complex. 

A poorer outcome is evident when using a Gamma distribution. Some of the heartbeats are correctly highlighted, but a good proportion of them can be confused with background likelihood. This is due to the long tail of the Gamma distribution, and thus, to the fact of applying a function defined on $\mathbb{R}^+$ to very short time series. Ultimately, this leads to an overestimation of the likelihood for those signal fragments overlapping every heartbeat.

%Figura de reconocimiento.

\subsection{Performance tests}
Table \ref{tab:performance} shows the computational time required to train PLHMM with different settings. Each row shows the results of learning an specific model from each heartbeat of Figure \ref{fig:heartbeat_morphologies}. The first column ('Discrete') shows the computational time in a scenario of unsupervised segmentation, after 4 iterations of training process, with a representation of duration given by a discrete probability distribution. The second column ('Discrete $[d^{min},d^{max}]$') shows the computational time in a scenario of semi-supervised segmentation, where physiological knowledge about common duration of different heartbeat waves and segments is provided as an interval $[d_j
^{min},d_j^{max}]$ for each state $S_j$. Thus, reestimation formulas are constrained by each interval, speeding up the training process. The third column ('Gamma') shows the computational time in a scenario of unsupervised segmentation, after 10 iterations of training process, and with a representation of duration given by the Gamma distribution. As already stated, training with the Gamma distribution requires an initialization, and mean values of each interval $[d_j^{min},d_j^{max}]$ were used for that purpose.

\begin{table}[h]

\begin{tabular}{llll}

\hline

Recording & Discrete & Discrete $[d^{min},d^{max}]$  & Gamma       \\ \hline

100  &  0:06:18.305  &  0:00:38.683  &  0:08:28.004 \\

103  &  0:04:35.944  &  0:00:26.137  &  0:06:26.389 \\

119  &  0:02:03.898  &  0:00:13.583  &  0:02:26.992 \\

106  &  0:04:21.194  &  0:00:30.800  &  0:05:48.052 \\ \hline

\end{tabular}
\caption{Performance results for training PLHMM.}
\label{tab:performance}
\end{table}

Performance tests were computed on a IBM-compatible PC, Intel\textsuperscript{\textregistered}
 Core\textsuperscript{TM} i5-7300HQ, CPU 2,50Ghz, RAM 8,00 GB.

\section{Conclusions and future work}
In this paper, a new model for one-shot learning of time series patterns is proposed. PLHMM is a sort of Hidden semi-Markov Model devised for representing a time series pattern as a sequence of regression probability distributions. Hence, unlike other proposals based on neural networks, PLHMM is easily interpretable. A method for learning each pattern from scratch has been designed, where just two initial parameters shall be provided: the number of latent states and the expressiveness of representation for each state, given by the number of basis functions. As it can be easily guessed, a lesser number of latent states entails a higher number of basis functions involved in their observable probability distributions, in order to warrant a good representation power. On the contrary, a higher number of latent states allows for a more simple representation of observable probability distributions, with a lesser number of basis functions. An optimal choice can be obtained by  exploring the use of information theory.

The first experiments on real time series have proven the potential of PLHMM to abstract a real pattern in the form of a probability distribution. In a real application to electrocardiography, the model shows an acceptable recognition ability. Further generalization can be achieved by designing a Bayesian updating of the model from new examples of the same category. 

On the other hand, even though an unsupervised learning strategy has provided good results, a semi-supervised one provide more interpretable results: in the realm of electrocardiography, using an a priori duration of the constituent waves of each heartbeat allows us to segment the time series according to a physiological meaning, and it will probably improve the stability of results. A complete validation against the MIT-BIH Arrhythmia Database should be performed to provide conclusive evidence for the claims supported by PLHMM.

As it can be seen, the Gamma distribution does not entail any improvement with respect to the discrete probability distribution, neither in terms of performance, nor in term of expressiveness. Different options of non parametric representations will be explored in the near future.

The main drawback of the model is the computational time it requires to be trained. To alleviate this issue, an interesting line of work would be to further explore initialization schemes based on previous learning results, following similar strategies as previously described in \cite{FeiFei06} . This may accelerate convergence, facilitating the use of PLHMM in the analysis of real data. Furthermore, performance improvement can be accomplished with parallel computing.

\section{Implementation}
With the aim of supporting reproducible research, the full source code of the algorithms presented in this paper has been published under an Open Source License\footnote{https://gitlab.citius.usc.es/adrian.perez/plhmm}, along with the ECG signal strips of all examples in this paper.

\end{document}